\documentclass{article}

\usepackage{arxiv}

\usepackage[utf8]{inputenc} 
\usepackage[T1]{fontenc}    
\usepackage{hyperref}       
\usepackage{url}            
\usepackage{booktabs}       
\usepackage{amsfonts}       
\usepackage{amsmath}        
\usepackage{nicefrac}       
\usepackage{microtype}      
\usepackage{lipsum}		
\usepackage{graphicx}
\usepackage{natbib}
\usepackage{doi}

\usepackage{float}
\usepackage{subfig}
\graphicspath{{images/}}
\usepackage{hyperref}
\usepackage{multirow} 

\usepackage{float}
\usepackage[table]{xcolor} 
\usepackage{array}
\usepackage[acronym]{glossaries}
\usepackage{xcolor}

\usepackage{subcaption} 

\title{A Modular Object Detection System for Humanoid Robots Using YOLO}

\date{September 13, 2025}	

\author{ 
    Nicolas Pottier\\
	School of Engineering and Computer Science\\
	Laurentian University\\
	\texttt{npottier@laurentian.ca} \\
	\And
    Meng Cheng Lau \\
	School of Engineering and Computer Science\\
	Laurentian University\\
	\texttt{mclau@laurentian.ca} \\
    \href{https://orcid.org/0000-0003-3517-4900}{\textbf{https://orcid.org/0000-0003-3517-4900}} \\
}

\renewcommand{\shorttitle}{}

\hypersetup{
pdftitle={A Modular Object Detection System for Humanoid Robots Using YOLO},
pdfsubject={cs.RO, cs.LG},
pdfauthor={Nicolas Pottier, Meng Cheng Lau},
pdfkeywords={},
}

\begin{document}
\maketitle

\begin{abstract}
Within the field of robotics, computer vision remains a significant barrier to progress, with many tasks hindered by inefficient vision systems. This research proposes a generalized vision module leveraging YOLOv9, a state-of-the-art framework optimized for computationally constrained environments like robots. The model is trained on a dataset tailored to the FIRA robotics Hurocup. A new vision module is implemented in ROS1 using a virtual environment to enable YOLO compatibility. Performance is evaluated using metrics such as frames per second (FPS) and Mean Average Precision (mAP). Performance is then compared to the existing geometric framework in static and dynamic contexts. The YOLO model achieved comparable precision at a higher computational cost then the geometric model, while providing improved robustness.
\end{abstract}

\begin{center}
\href{https://sites.google.com/laurentian.ca/limrl}{\textbf{Laurentian Intelligent Mobile Robotic Lab}}
\end{center}



\section{Introduction}
In computer vision, AI (Artificial Intelligence) powered frameworks have significantly increased performance in both speed and accuracy. Conventional object detection strategies have been rendered increasingly obsolete by object detection AI, such as CNNs (Convolutional Neural Networks). In robotics, applications of CNNs have remained limited due to their large computational cost on constrained systems. Machine learning frameworks typically require large datasets and a powerful GPU (Graphical Processing Unit) to achieve good performance. Robotic environments rarely have access to large amounts of memory, and their GPUs tend to be limited, or even non-existent. Recently, CNN frameworks have seen substantial performance improvements. Specifically, the latest YOLO (You Only Look Once) model, YOLOv9, outperforms almost every other framework in most aspects \cite{wang2024yolov9learningwantlearn}. 

This paper presents the design and optimization of a lightweight, general-purpose vision module optimized for human robotics using YOLOv9. To evaluate its utility, we examine performance within the context of the FIRA (Federation of International Robot-Sports Association) Hurocup, a competition in which robots are tasked with performing visually intensive tasks such as archery, weightlifting, and basketball. The vision module is designed to run on a CPU-only platform, reflecting the limitations commonly found in humanoid robots

Our objectives are threefold : (1) assess whether YOLO-based detection can operate efficiently under the hardware constraints typical of humanoid robots; (2) compare its performance to traditional vision techniques; and (3) discuss the broader implications of integrating object detection frameworks in robotics. By demonstrating the viability of real-time YOLO-based detection on constrained systems, this work contributes to closing the gap between cutting-edge computer vision research and its deployments in field robotics.


\section{Literature Review}

\subsection{Object Detection}
Computer vision encompasses many subfields, with object detection being a critical area due to its growing relevance in real-time applications such as autonomous vehicles \cite{8621614}, agriculture \cite{wang2020real,salazargomez2021practicalobjectdetectionweed,KHAN2022108201}, surveillance \cite{murugan2018study,angadi2020review}, and healthcare \cite{ahmed2025deep,WANG2022103945}. Modern object detection involves both classifying objects and localizing them with bounding boxes—tasks essential for systems requiring fast, accurate decisions.

Early methods relied on feature-based techniques like Histogram of Oriented Gradients (HOG), which showed high accuracy in pedestrian detection \cite{suard2006pedestrian} and facial recognition \cite{DENIZ20111598}. Scale-Invariant Feature Transform (SIFT) improved matching through keypoint detection \cite{doi:10.1504/IJISC.2020.104826,s130709248}, while Speeded-Up Robust Features (SURF) enhanced performance for real-time tasks \cite{LI20122094,he2019multimedia}.

The rise of deep learning and convolutional neural networks (CNNs) revolutionized object detection. Two-Stage Detectors like Fast R-CNN emphasize accuracy and have shown success in pavement crack detection \cite{s22031215} and facial expression recognition \cite{LI2017135}. One-Stage Detectors, including SSD and the YOLO family, prioritize speed, making them ideal for real-time use.

While AI-powered vision has been widely applied in areas like surgical robotics \cite{AI2023105478} and infrastructure monitoring \cite{WARD20211253}, humanoid robotics lags behind. Most research in that domain emphasizes interaction and learning \cite{1014810,1442688,BAO20231027,5509181}, rather than detection. This paper addresses that gap by developing a robust vision module suited for humanoid platforms, leveraging recent advances in real-time object detection.

\subsection{YOLO: You Only Look Once}
The YOLO framework is a leading one-stage object detector used in both research and industry \cite{drones7030190,machines11070677}. As a CNN model, YOLO processes images through convolutional, pooling, and fully connected layers to detect and localize objects with labeled bounding boxes and confidence scores.

YOLOv1, introduced in 2015, pioneered the single-stage detection approach and achieved 45 FPS with 63.4\% average precision (AP) \cite{redmon2016lookonceunifiedrealtime}. YOLOv2 followed with major improvements, reaching 78.6\% AP at 40 FPS \cite{redmon2016yolo9000betterfasterstronger}, and YOLOv3 further refined the model in 2018 \cite{redmon2018yolov3incrementalimprovement}.

YOLOv4 (2020) introduced key techniques such as mosaic augmentation and drop block regularization \cite{bochkovskiy2020yolov4optimalspeedaccuracy}. Subsequent versions, including YOLOv5 and YOLOv6, enhanced performance further; YOLOv6 notably achieved 35.9\% AP at 1234 FPS \cite{li2022yolov6singlestageobjectdetection}. YOLOv7 improved both speed and accuracy \cite{wang2022yolov7trainablebagoffreebiessets}.

The latest version, YOLOv9, introduces programmable gradient information and the GELAN architecture, pushing the framework to new performance heights. Though widely adopted across domains, YOLOv9 has yet to be implemented on a humanoid robot.

\subsection{ Robotic Applications \& Vision Modules}
Vision modules are critical to robotic systems, serving as the interface between sensory input and decision-making. Their primary function is to interpret visual data for tasks such as navigation and object detection. Earlier systems relied on classical techniques like segmentation, edge detection, and template matching. While effective in some cases, these methods struggle under real-world variability and noise. Recent advances in AI have greatly improved the adaptability and accuracy of vision modules, making research in this area crucial for modern robotics.

Classical approaches—such as color segmentation and template matching—were once standard for robotic vision \cite{https://doi.org/10.1155/2018/1969834} \cite{HURTADO2022279}. These methods are fast and require little processing power, but they lack robustness. Variability in lighting, background clutter, and object similarity can easily confuse such systems. For instance, similarly colored objects or reflective surfaces can lead to detection failures and hinder performance.

The emergence of deep learning revolutionized robotic vision. CNN-based frameworks like YOLO, Faster R-CNN, and SSD (Single-shot Detectors) allow for more precise detection in complex scenes \cite{redmon2016lookonceunifiedrealtime} \cite{Liu_2016} \cite{ren2016fasterrcnnrealtimeobject}. These models, trained on large datasets, generalize far better than traditional methods. However, their implementation on robots introduces new challenges. Most robots lack GPUs, making real-time CPU inference difficult. Additionally, building large, diverse training datasets is time-consuming and resource-intensive. Variations in lighting, occlusion, and clutter still pose issues \cite{doi:10.1504/IJISC.2020.104826}. Designing vision modules for humanoid robots thus requires carefully balancing accuracy, speed, and hardware constraints.

\subsection{Research Gap}
This research serves to address a critical gap in the intersection of Computer Vision, YOLO-based frameworks, and robotic vision. While YOLO has been extensively used for tasks such as autonomous driving and surveillance, its applications in humanoid robotics remain very limited. Current implementations of vision modules rely on task-specific configurations and traditional techniques. This can lead to struggles in tasks where a more generalized approach is required or in unpredictable environments such as FIRA competitions. This paper leverages YOLO to create a generalized vision module that balances efficiency and accuracy, enabling its deployment onto resource-constrained humanoid robots. This will contribute to the field by addressing the dual challenge of computational constraints and the lack of general-purpose vision solutions for humanoid robots. By employing various optimization tools and making the most of performance improvements, the YOLO framework might finally be accessible for humanoid robotics.

\section{Methodology}

\subsection{Hardware Configuration}

\begin{figure} 
    \centering
    \includegraphics[width=0.6\textwidth]{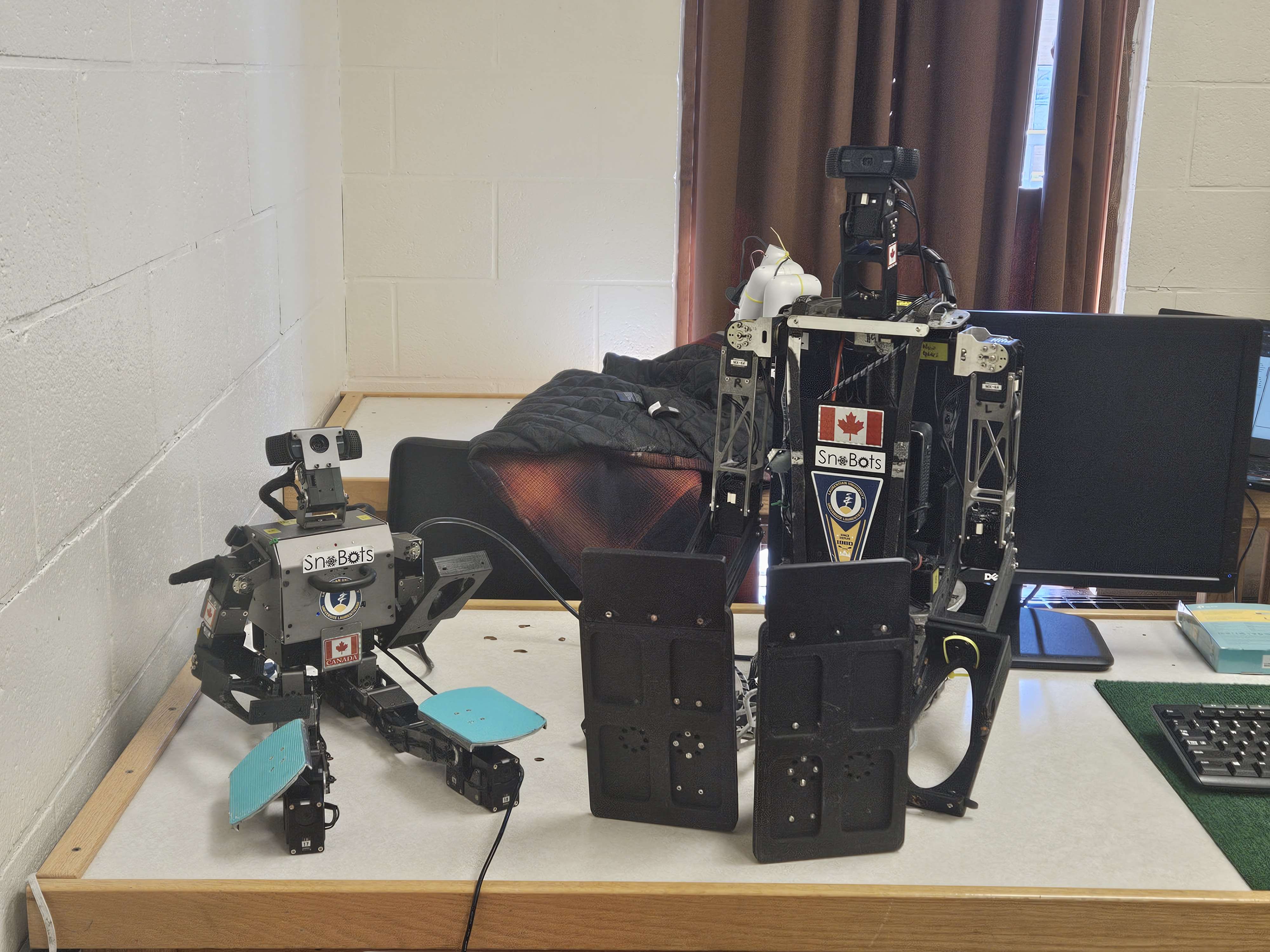}
    \caption{ LIMRL's two humanoid robots. ROBOTIS-OP3, also known as Oscar (pictured left), will be used for this implementation. The vision module will also be installed on Polaris (pictured right).}
    \label{fig:OSCAR}
\end{figure}

The humanoid robot used in this paper is the ROBOTIS-OP3 (See Fig. \ref{fig:OSCAR}), a 51 cm tall, 3.5 kg platform equipped with 20 XM430-W350-R actuators and a Logitech C920 HD Pro Webcam. It’s powered by an 11.1V 1800mA LiPo battery or a direct cable connection. Minor modifications include 3D-printed plastic hands and feet. The OP3 runs ROS1 Kinetic, enabling smooth hardware interaction, while its computational efficiency and real-time processing make it suitable for object detection tasks. Local image processing reduces latency, though its modular architecture allows easy transition to external computing if needed. Connections can be made via a VNC client or an Ethernet cable.

\subsection{YOLOv9 Preparation and Training}

As previously outlined, this project integrates the YOLO object detection framework into a humanoid robot to enable real-time detection within hardware constraints and ensure fast communication between vision and control systems. The software stack combines pre-trained models, open-source libraries, and middleware to maintain both flexibility and reproducibility. The implementation is built on the Robot Operating System (ROS) for its modular design and rich library support. ROS enables seamless integration of the camera, YOLO inference, and actuator control. Python is used for development, leveraging PyTorch and OpenCV—both essential for YOLO-based computer vision tasks.

YOLO was chosen for its strong balance between speed and accuracy. The latest version, YOLOv9, introduces Programmable Gradient Information (PGI) and the GELAN architecture, enhancing inference time and accuracy, making it well-suited for robotic deployment. This paper applies YOLO to three FIRA Hurocup events: Archery, Marathon, and Basketball. Table \ref{tab:yolo_obj} shows an example of the objects used in training. In Archery, the model detects a moving target and fires an arrow at its center. In Marathon, the robot follows a red line and classifies directional arrow markers (left, right, forward), which may appear rotated. For Basketball, the robot detects an orange ball, retrieves it, locates a red net, approaches it, and shoots when close enough.

\begin{table}
    \centering
    \renewcommand{\arraystretch}{1.2} 
    \setlength{\tabcolsep}{6pt} 
    \caption{Table describing which objects will be included within the datasets, organized by event.}
    \begin{tabular}{|m{3cm}|m{7cm}|m{3cm}|} 
        \hline
        \rowcolor{green!30}
        \multicolumn{3}{|c|}{\textbf{YOLOv9 Training Object Examples}} \\ 
        \hline
        \textbf{Model Name} & \textbf{Description} & \textbf{Image} \\
        \hline
        \rowcolor{gray!30}
        \multicolumn{3}{|c|}{\textbf{Basketball}} \\
        \hline
        Ball & A ping pong ball is used for the kid size, and a larger tennis ball for the adult size. & 
        \makebox[3cm]{\includegraphics[width=2cm]{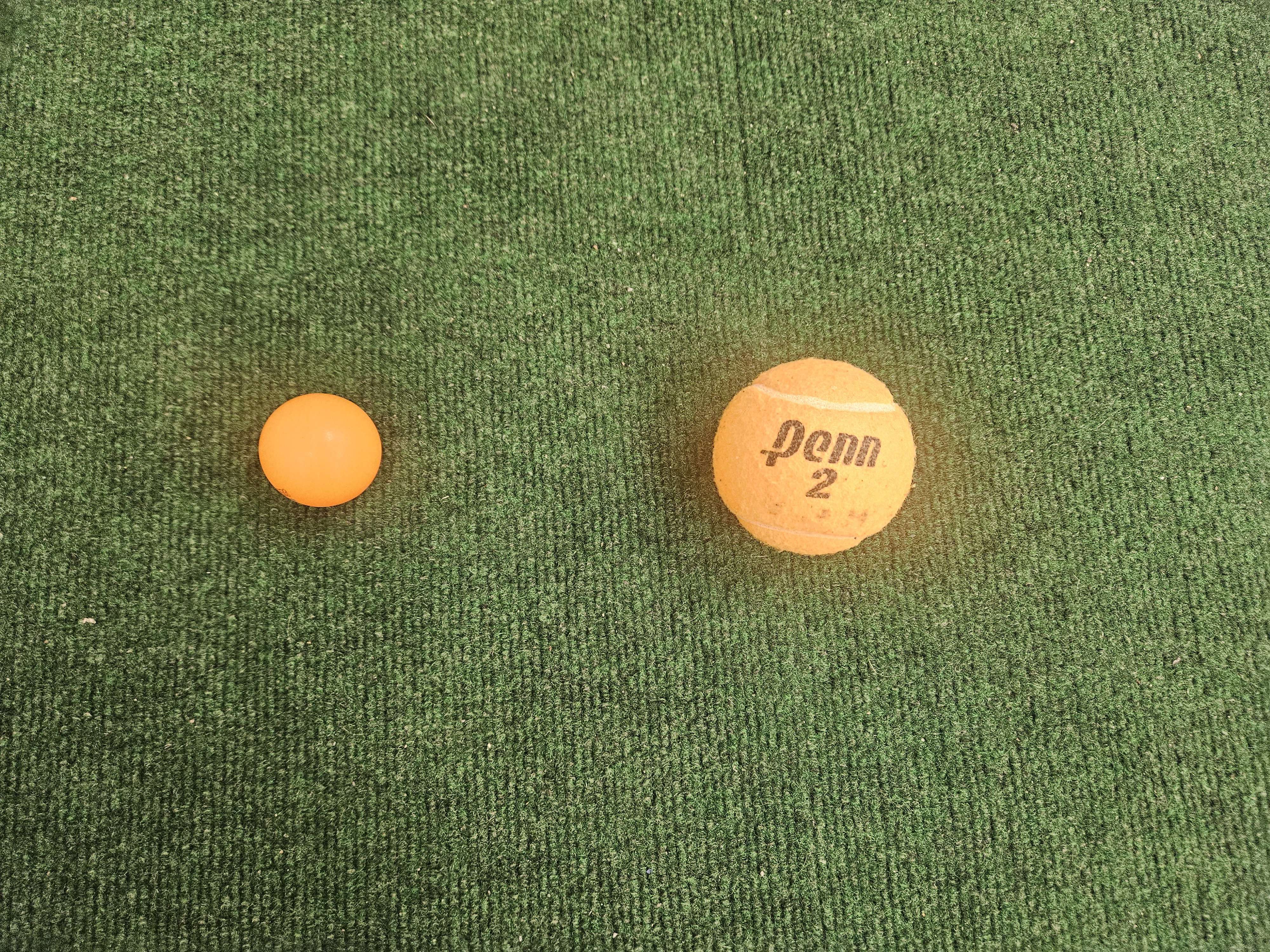}}  \\
        \hline
        Basket & A bright basketball where the robot must be able to score points on. The kid-sized basket has a diameter of 10cm, while the adult-sized basket has a diameter of 30cm. & 
        \makebox[3cm]{\includegraphics[width=2cm]{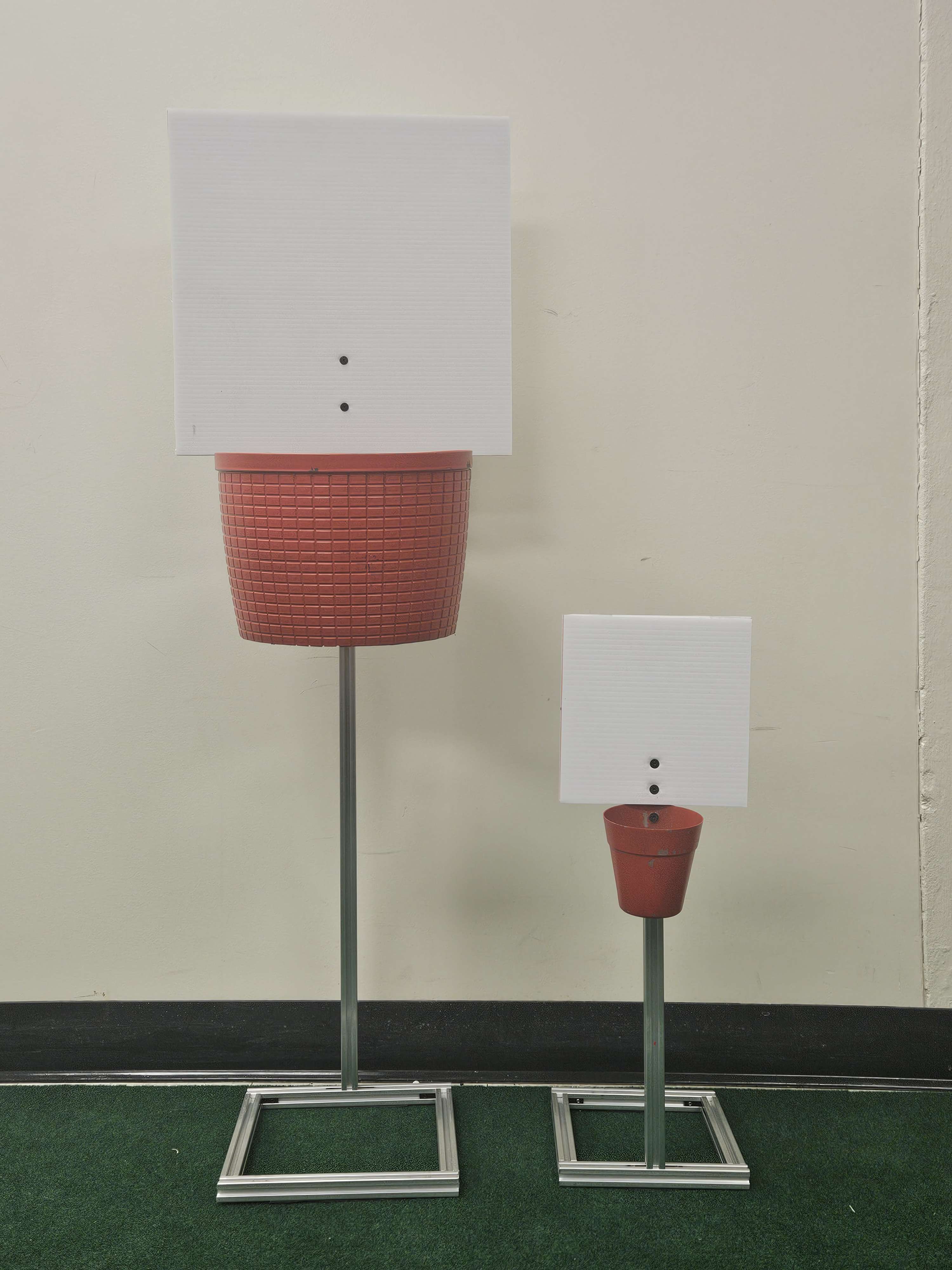}} \\        
        \hline
        \rowcolor{gray!30}
        \multicolumn{3}{|c|}{\textbf{Archery}} \\
        \hline
        Target & An archery target with a diameter of 50cm. & 
        \makebox[3cm]{\includegraphics[width=2cm]{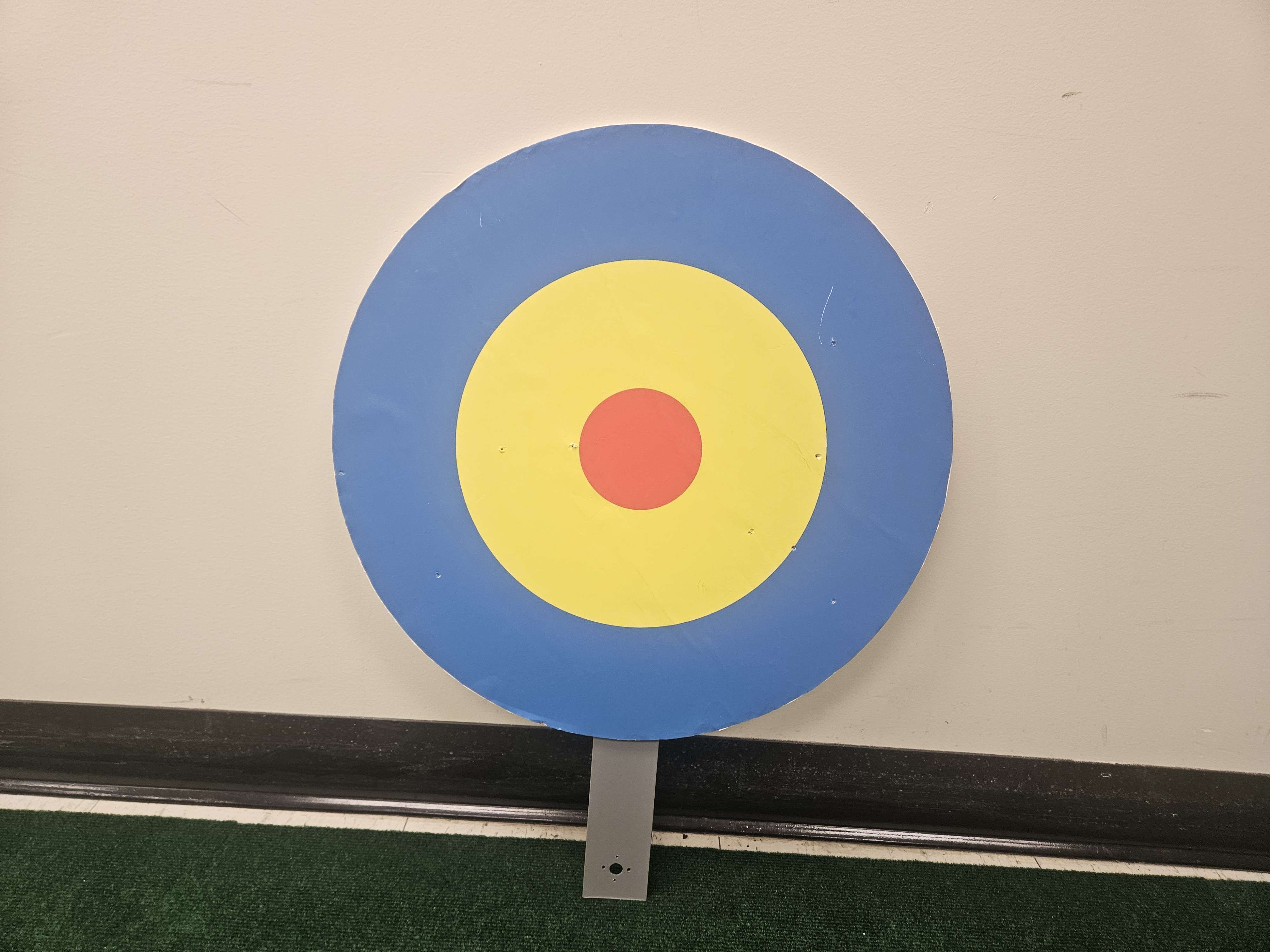}}  \\
        \hline
        \rowcolor{gray!30}
        \multicolumn{3}{|c|}{\textbf{Marathon}} \\
        \hline
        Line & A red guiding line which the robot is tasked with following. & 
        \makebox[3cm]{\includegraphics[width=2cm]{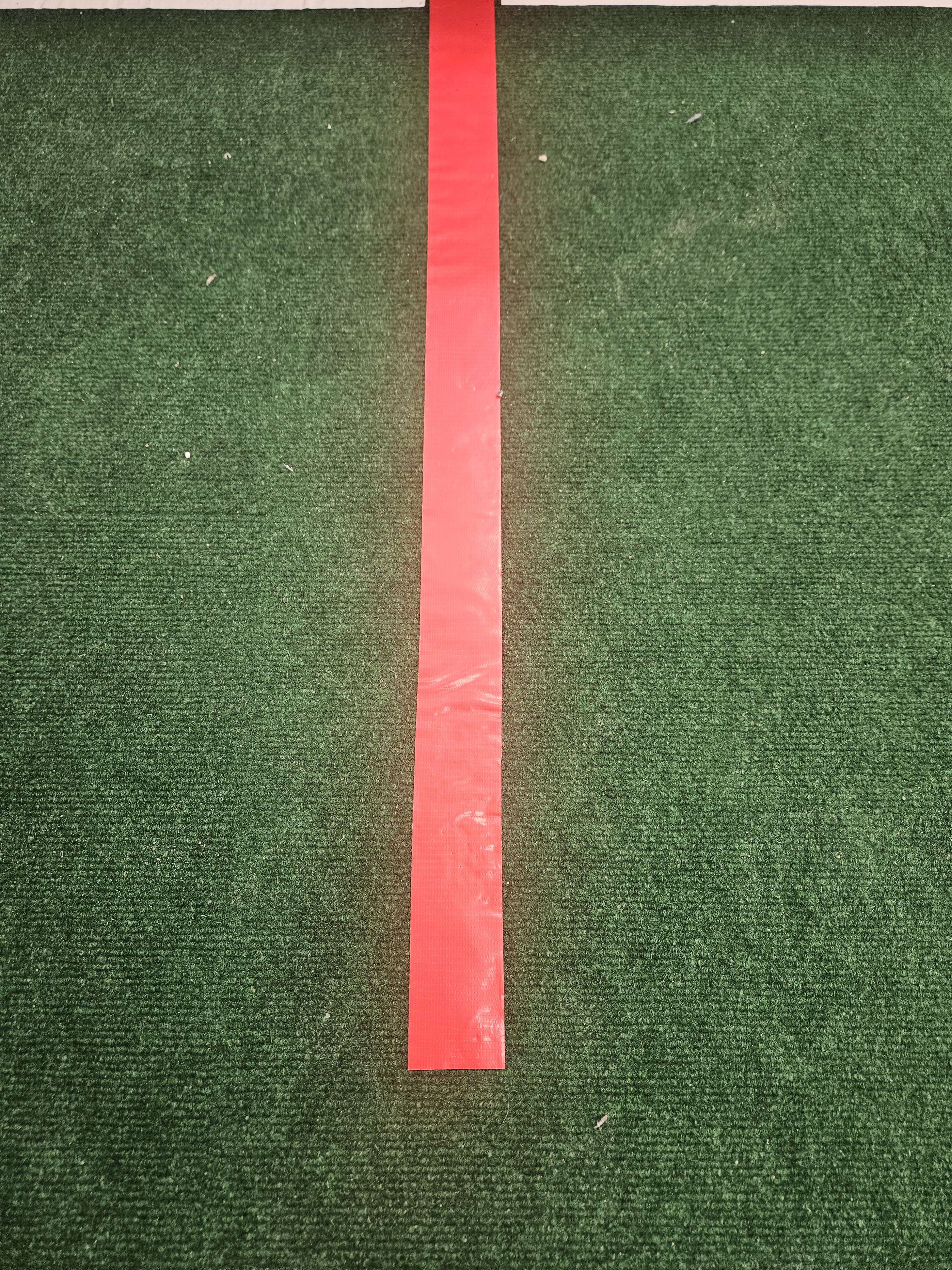}}  \\
        \hline
        Right Arrow & An arrow whose tip points to the right. & 
        \makebox[3cm]{\includegraphics[width=2cm]{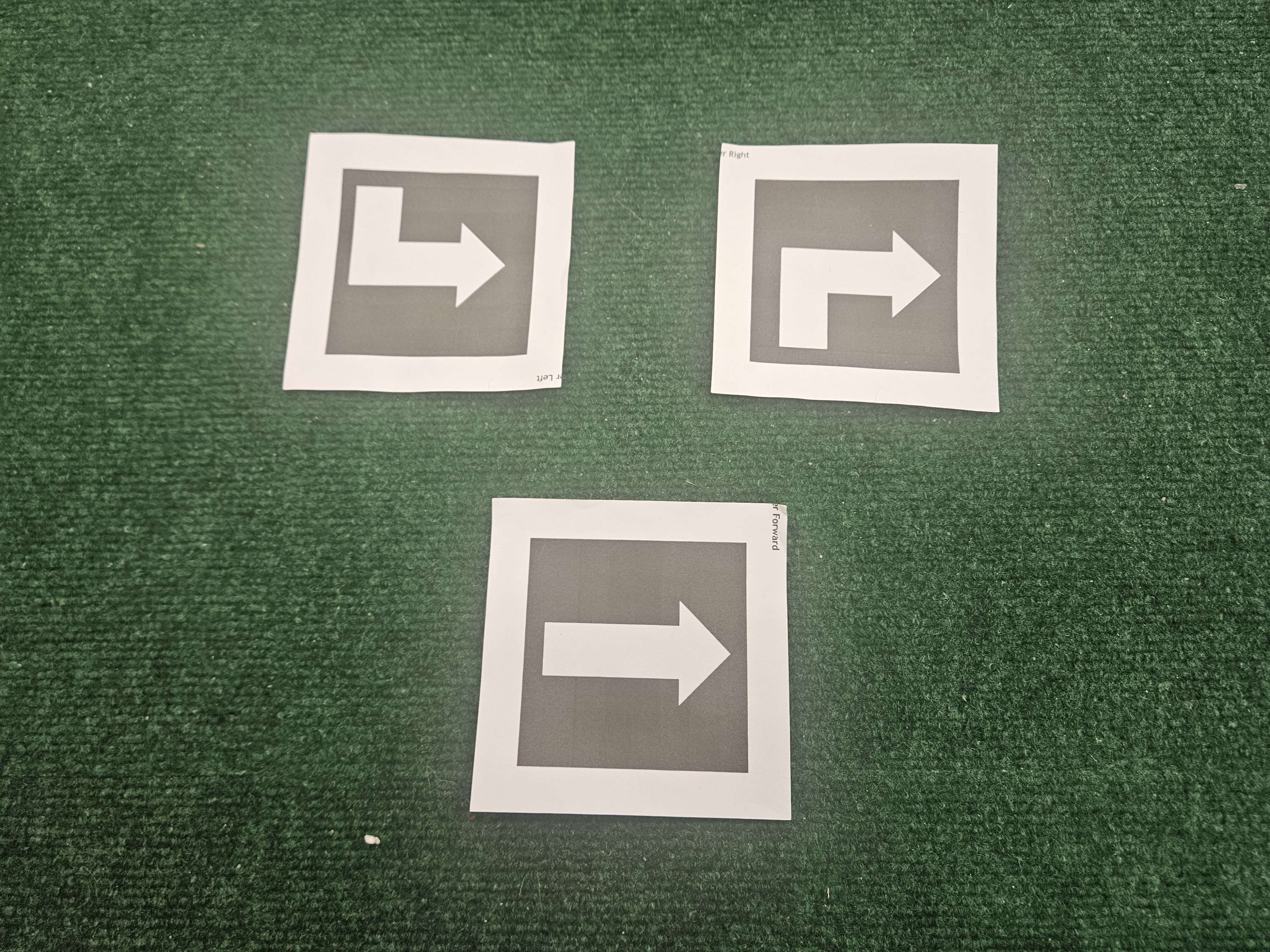}}  \\
        \hline
        Left Arrow & An arrow whose tip points to the left. & 
        \makebox[3cm]{\includegraphics[width=2cm]{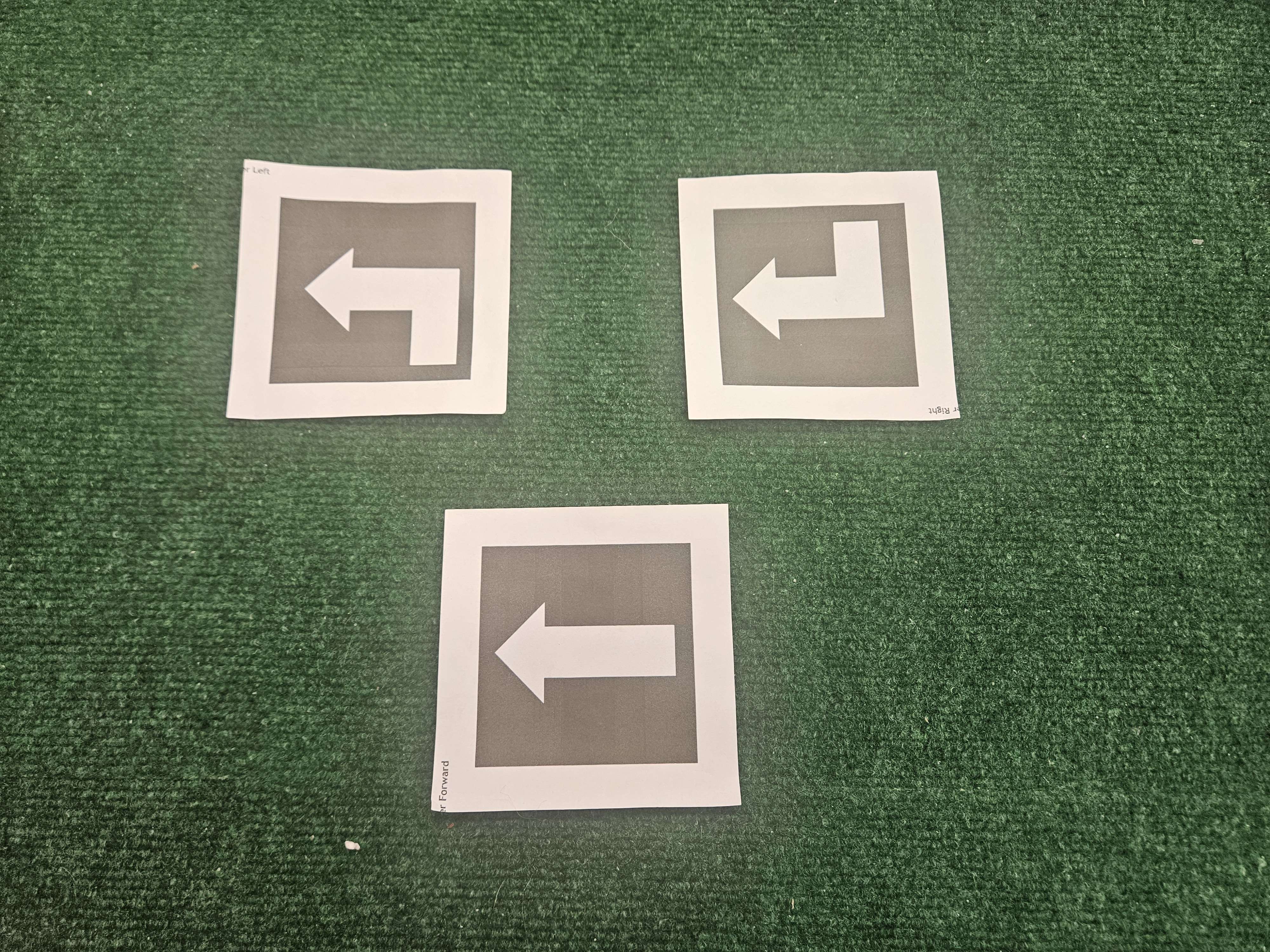}} \\
        \hline
        Forward Arrow & An arrow whose tip points straight ahead. & 
        \makebox[3cm]{\includegraphics[width=2cm]{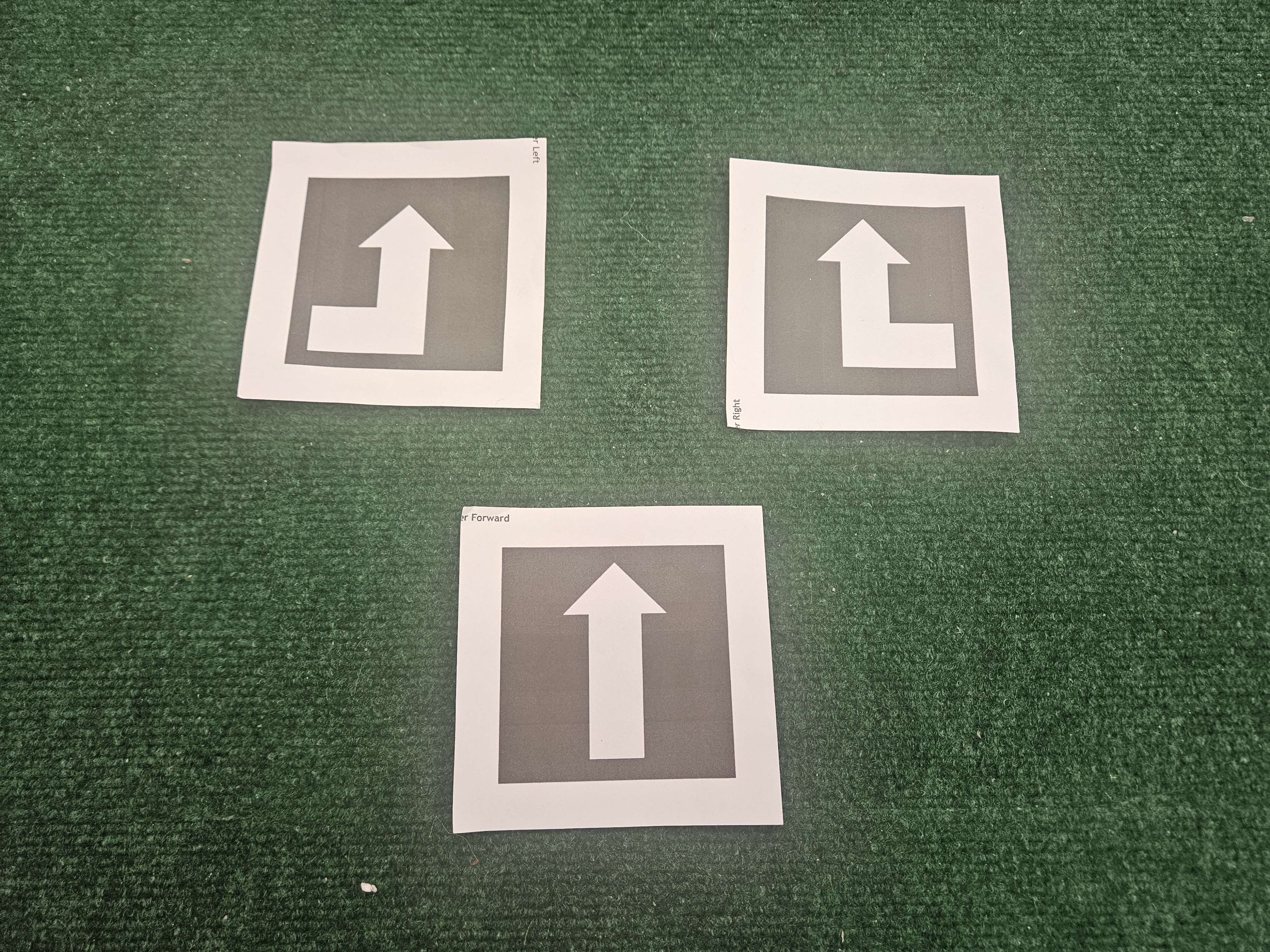}} \\
        \hline
        \rowcolor{gray!30}        
    \end{tabular}
    \label{tab:yolo_obj}
\end{table}

The training process begins with collecting and preprocessing task-specific datasets. To maintain performance and efficiency, multiple smaller models are trained for individual tasks rather than a single large one. Each dataset corresponds to specific FIRA Hurocup events—for example, balls and nets for basketball or arrow markers for the marathon (see Table 3.1). Datasets are annotated using bounding boxes in Roboflow's YOLOv9 format, where labels include class ID and normalized bounding box coordinates. Roboflow also handles augmentation tasks—such as rotation, lighting changes, and scaling—to improve generalization \cite{10.1145/3678935.3678956,10420361,10560330}.

Training was conducted using PyTorch on an Nvidia RTX 1650 GPU. Models were trained over 100 epochs with a batch size of six and a close mosaic threshold of 50. To avoid overfitting, a patience flag of five was set, halting training if no improvement occurred within five epochs—an optimal value determined through prior experimentation. Images were processed at 640×640 resolution, and training began from pre-trained MSCOCO weights \cite{ultralytics_yolov9}. Three architectures were used: YOLOv9-Tiny, YOLOv9-Small, and YOLOv9-Medium—the most lightweight variants. Training was executed via command line on an external laptop to reduce onboard computational load.

\subsection{ROS Integration}
The vision module integrates tightly with the Robot Operating System (ROS), ensuring seamless communication between the camera, inference pipeline, and OP3's actuators. The architecture consists of three core components: the YOLO node, a listener node, and event-specific nodes.

ROS uses a publisher/subscriber communication model, allowing nodes to exchange data asynchronously through named topics. This modular approach enables system flexibility—nodes can operate independently while exchanging data efficiently. In this system, the YOLO node publishes detection results to a topic, which the listener node subscribes to, formats, and routes to the appropriate event modules.

\begin{figure} 
    \centering
    \includegraphics[width=0.8\textwidth]{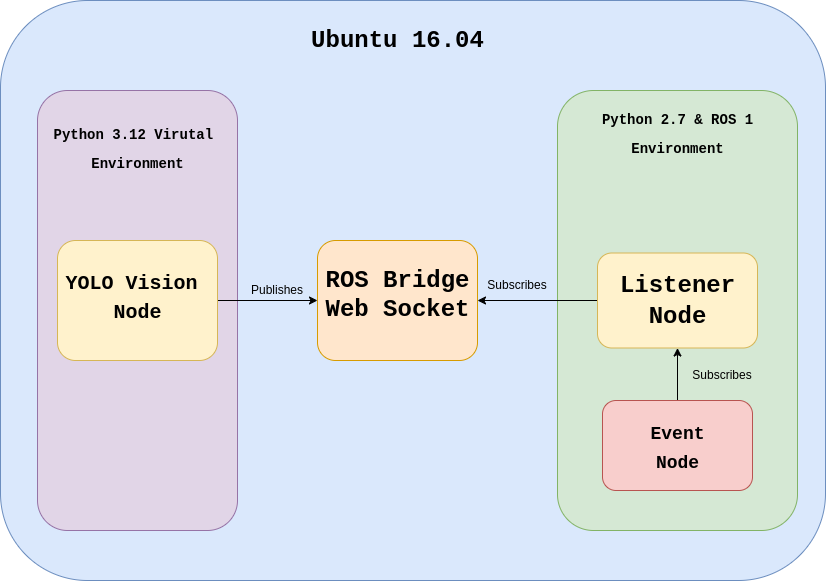}
    \caption{Flowchart showing the interaction between ROS Nodes within the vision model's various environments. The ROS Bridge WebSocket is essential to ensure that both environments can communicate properly.}
    \label{fig:Env}
\end{figure}

Due to compatibility issues—ROS relying on Python 2.7, OpenCV 3.3.1, and YOLO requiring Python 3—the YOLO node is run in a Python 3.12 virtual environment with OpenCV 4.11. This isolation allows it to leverage optimized libraries like PyTorch for model inference. The node subscribes to a camera feed, processes video frames resized to the model’s input dimensions, and performs object detection. Detections (bounding boxes, labels, and confidence scores) are published via WebSocket as JSON. The listener node then receives this data and republishes it on ROS topics suitable for each event. To optimize performance, inference can be limited to 20 detections per second rather than processing every frame. Fig. \ref{fig:Env} provides a visual representation of the various environments used.

The aforementioned issues between YOLO and Python 2.7 introduce the need for a separate listener node to receive detection data via WebSocket and republish it to the /yolo/bboxes topic. This node enables event-specific modules to access detection results in a ROS-compatible format without direct integration with the YOLO node. Each event node subscribes to /yolo/detections and uses the data to identify targets and initiate actions such as approach or avoidance. Object tracking is handled by pre-written Python logic, while actuator control is managed through the Dynamixel SDK.

Given the limited CPU power onboard OP3, real-time inference performance remains a challenge. To mitigate this, training is performed externally, while onboard inference is optimized using frame skipping and resizing inputs to 640×640 resolution. Since OP3 lacks a GPU, the YOLO model is converted to OpenVINO for CPU efficiency. This involves first converting the model to ONNX format, where quantization reduces computational load. The modular nature of ROS further allows for easy substitution or scaling with alternative detection models.

\subsection{Performance Evaluation}

To ensure reliable performance, the new vision module will undergo comprehensive testing in both static and dynamic environments. Initial evaluations will take place in a controlled laboratory setting with minimal background interference and fixed lighting. From a stationary position, the robot’s ability to detect objects will be assessed in order to measure frame rate and latency in a reproducible environment.

Following static tests, the vision module will be evaluated in real-world conditions by completing three FIRA Hurocup events: Archery, Basketball, and Marathon. Performance will be assessed using quantitative metrics such as FPS, Precision, Recall, mAP50, and mAP50–95. Static and dynamic FPS measurements will help evaluate latency and consistency. Distance estimation will be tested by comparing the vision module’s output to a geometric approach and ground truth measurements. This comparison will highlight the system’s accuracy and its advantages over previous methods. Extended testing will ensure robustness and guide iterative improvements to detection accuracy in dynamic conditions.

\section{Implementation}

\subsection{Model Deployment}

The deployment of the YOLO model involved a lengthy training process, beginning with dataset preparation and culminating in real-time inference. The dataset was composed of no more than 150 images per class to attain a balance between speed and accuracy. The data sets were manually labeled using RoboFlow to help the model focus on key objects, such as basketballs or arrow markers. Data augmentation techniques such as rotation, shear, exposure adjustment, and Gaussian blur were applied to improve overall robustness. 

For real-time inference, a video feed is captured through the ROBOTIS-OP3's webcam. Each captured frame was processed using a letterbox resizing function to maintain the image's aspect ratio while fitting the required input size (640px x 640px). Several techniques were utilized to help optimize the model for a CPU-based inference, including model conversion into OpenVINO format. The bounding boxes are drawn only if they are over a confidence threshold, and non-max suppression (NMS) is applied to filter overlapping bounding boxes, retaining only the most accurate detections. Bounding boxes were then scaled back to the original image dimensions, correcting for the webcam's aspect ratio and avoiding distorted bounding boxes. This step was crucial to ensure accurate placement of detections in the video feed. OpenCV 4.11 was used to draw bounding boxes and display class labels on the detected objects, along with a confidence score. 

\subsection{ROS Node Developpement}

The YOLO object detection system is implemented using two separate ROS nodes due to Python version incompatibilities. ROS Kinetic requires Python 2.7, whereas YOLOv9 and its dependencies (e.g., PyTorch) need Python 3.8 or newer. To resolve this, the detection node runs in a Python 3.12 virtual environment, isolated from ROS. A second node—the listener node—operates in Python 2.7 and receives the detection results via WebSocket through a ROS bridge. This bridge enables seamless communication between the two environments.

The detection node performs inference using PyTorch and Ultralytics' YOLOv9, sending bounding box data as JSON messages over WebSocket. Each JSON includes object class, coordinates, size, and confidence score. OpenCV 4.11 is used to overlay detections on the video feed for debugging. The node also includes a cleanup routine to release camera resources and close WebSocket connections on shutdown.

The listener node receives the JSON data and republishes it as ROS-compatible messages. Running within the ROS environment, it ensures that all other event-specific nodes can subscribe to detection outputs without needing to interact directly with YOLO. The node logs incoming data to rospy for debugging and includes error handling for malformed messages or connection failures. WebSocket lifecycle events are managed to allow reconnections and diagnostic logging.

This dual-node architecture ensures modern deep learning tools can be used without disrupting ROS compatibility. The modular design promotes maintainability and scalability, allowing future upgrades to the detection pipeline without altering the core ROS control structure.

\subsection{Pipeline and Behavior Integration}

The new vision module is integrated into the robot’s framework using a structured pipeline built on ROS architecture. This setup enables efficient data exchange between the vision system and behavioral modules, particularly for FIRA events, allowing real-time object detection to drive robot actions.

The main implementation challenge stemmed from YOLO's Python 3 requirement conflicting with ROS Kinetic's Python 2.7. To solve this, a WebSocket bridge was created to connect a Python 3.12 virtual environment (hosting the YOLO node) with the ROS system. Initial setup issues—such as the virtual environment defaulting to Python 2.7 libraries like OpenCV 3.3.1 and PIP—were resolved by restructuring the environment with the --no-site-packages flag. This allowed the isolated vision node to communicate with the ROS system without dependency conflicts.

The YOLO vision node sends detection data as JSON messages over WebSocket. These are received by a listener node, which repackages the data into ROS-compatible messages. This decoupled, asynchronous communication ensures low-latency performance—critical for the fast reactions required in FIRA events.

Event nodes can now easily subscribe to the updated detection topics. Launching an event is as simple as updating the roslaunch file to point to the new vision node. Once active, these nodes interpret bounding box data to determine object location and trigger behaviors such as tracking, turning, or avoidance, depending on the object's position and proximity.

\begin{figure}
    \centering
    \includegraphics[width=1.05\linewidth]{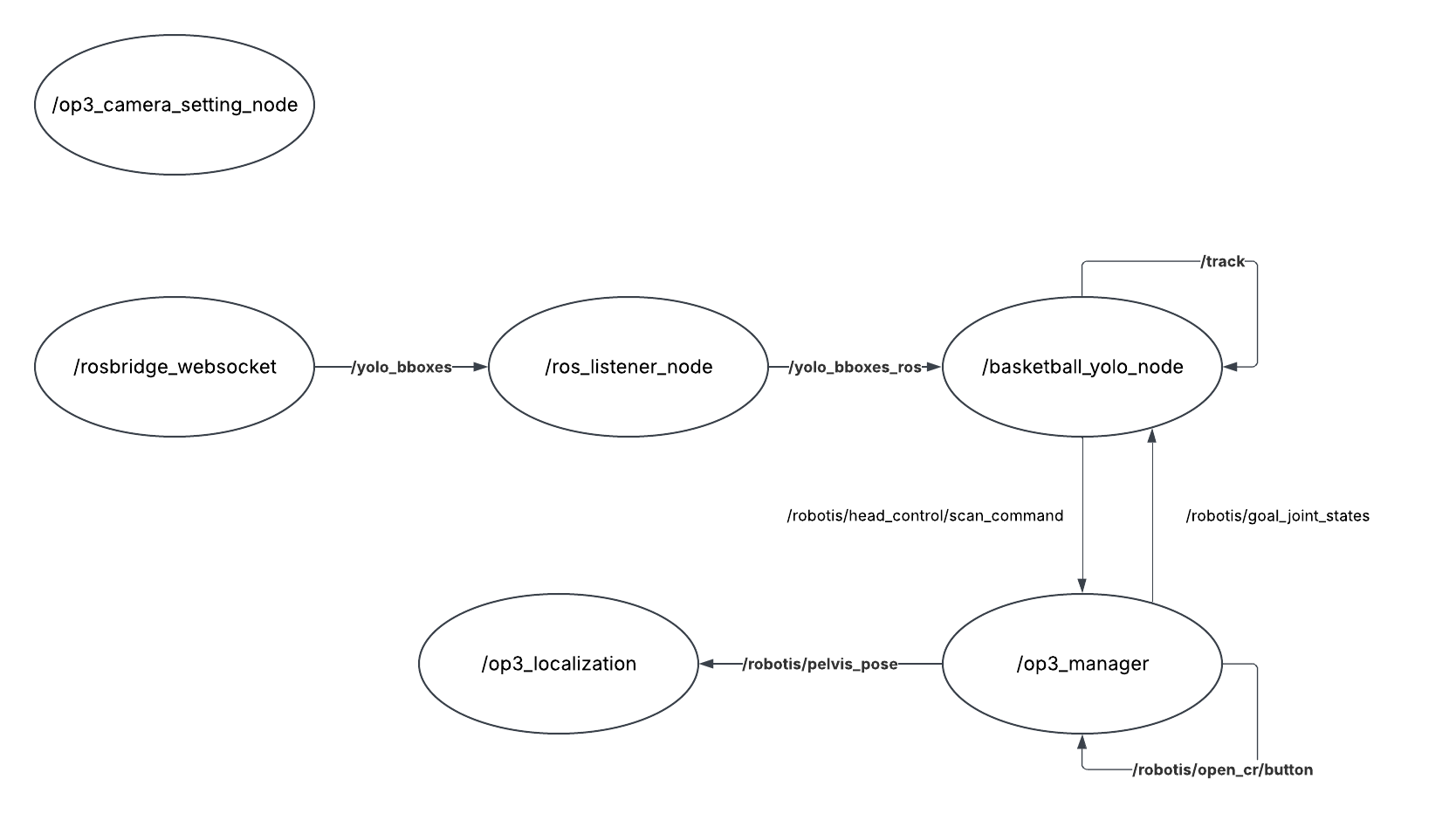}
    \caption{Graph showing the pipeline utilized by Event scripts. The graph shown is for the basketball event.}
    \label{fig:rosg}
\end{figure}

The ROS publisher-subscriber model ensures each component processes only the data it needs, improving efficiency and scalability. This modular structure preserves ROS core stability while supporting extensibility, making it ideal for future upgrades. The separation of perception from decision-making enhances maintainability and ensures the system can support additional tasks beyond the FIRA competition. An example of the complete pipeline during an event can be found in Fig. \ref{fig:rosg}

\subsection{ Optimization }
Optimizing the YOLOv9 model is essential to achieving real-time performance on the ROBOTIS-OP3’s CPU-only system. Several strategies—frame skipping, input resizing, model conversion, and quantization—were employed to reduce inference latency while preserving detection accuracy.

One of the most effective methods was frame skipping. Rather than processing every incoming frame, the system processes every nth frame, where n is a configurable parameter. This significantly reduces computational load while maintaining responsiveness, allowing the robot to track and react to objects without overloading the CPU. Another crucial optimization was input resizing. Camera frames are downscaled to 640×640 pixels—the native YOLO input size—prior to inference. This ensures efficient processing without unnecessary high-resolution computations, speeding up preprocessing and inference while retaining detection reliability. To further reduce inference time, the YOLOv9 model was converted from PyTorch to ONNX (Open Neural Netowrk Exchange), and then from ONNX to OpenVINO, Intel’s toolkit optimized for CPU performance. The ONNX conversion ensures model portability, while OpenVINO accelerates inference by leveraging CPU-specific instructions and pruning redundant operations. Finally, quantization was applied to reduce model size and inference time. PyTorch’s quantization tools lowered weights and activations from 32-bit floats to 8-bit integers.This step greatly decreased memory usage and sped up matrix operations—the core bottleneck of inference.

Together, these optimizations enable the YOLOv9 system to function efficiently within OP3’s hardware limitations. By combining model-level and hardware-specific techniques, the system strikes a strong balance between speed and accuracy, allowing real-time object detection in demanding robotic environments.

\section{Results}
\subsection{YOLO Training Results}

Training the YOLOv9 models was a crucial step in ensuring high detection accuracy while maintaining real-time performance. For this implementation, nine models were trained—three per FIRA event—using the YOLOv9-Tiny (T), Small (S), and Medium (M) architectures. This multi-model strategy allowed for a balance between accuracy and speed, essential for deployment on the limited hardware of the ROBOTIS-OP3. To improve model accuracy, all models employed weights that were pre-trained on the MSCOCO dataset. These weights can be found at \cite{ultralytics_yolov9}. Additionally, all models were trained with a patience of 5, as to avoid overfitting. Results for the training can be found in Table \ref{tab:train}

\begin{table}[h]
    \centering
    \renewcommand{\arraystretch}{1.5} 
    \setlength{\tabcolsep}{10pt} 
     \caption{Results of YOLO model training. All models were trained on 100 epochs with a patience factor of 5}
    \begin{tabular}{|c|c|c|c|c|}
        \hline
        \rowcolor{orange!50}
        \multicolumn{5}{|c|}{\textbf{YOLOv9 Training Results}} \\ 
        \hline
        \textbf{Model Name} & \textbf{mAP@0.5} & \textbf{mAP@0.5-0.95} &\textbf{Precision} &\textbf{Recall} \\
        \hline
        \rowcolor{gray!30}
        \multicolumn{5}{|c|}{\textbf{Basketball}} \\
        \hline
        YOLOv9-T & 0.994 & 0.92536 & 0.98565 & 0.98383 \\
        YOLOv9-S & 0.995 & 0.96225 & 0.98253 & 0.99809 \\
        YOLOv9-M & 0.99472 & 0.95518 & 0.9859 & 0.98571\\
        \hline
        \rowcolor{gray!30}
        \multicolumn{5}{|c|}{\textbf{Archery}} \\
        \hline
        YOLOv9-T & 0.95946 & 0.92109 & 0.95934 & 1 \\
        YOLOv9-S & 0.98321 & 0.91407 & 0.95782 & 1\\
        YOLOv9-M & 0.99393 & 0.82793 & 0.95612 & 1\\
        \hline
        \rowcolor{gray!30}
        \multicolumn{5}{|c|}{\textbf{Marathon}} \\
        \hline
        YOLOv9-T & 0.75063 & 0.68905  & 0.67078 & 0.89904 \\
        YOLOv9-S & 0.78545 & 0.71538  & 0.61382 & 1\\
        YOLOv9-M & 0.73501 & 0.61436 & 0.5923 & 0.87498\\
        \hline
    \end{tabular}
    \label{tab:train}
\end{table}

For this implementation, nine models were trained—three per FIRA event—using the YOLOv9-Tiny (T), Small (S), and Medium (M) architectures. This multi-model strategy allowed for a balance between accuracy and speed, essential for deployment on the limited hardware of the ROBOTIS-OP3.

Training results varied by event. In the Basketball event, all three models achieved high scores across mAP, precision, and recall metrics, indicating strong generalization and reliable detection across frames.

In contrast, Archery models exhibited mixed results. Despite high recall and mAP0.5, the mAP0.5–0.95 scores varied significantly. This, along with elevated recall scores, suggests overfitting: the models successfully identify targets seen during training but fail to generalize well to new data. Further investigation or dataset augmentation may be necessary to improve their robustness.

\begin{figure} [h]
    \centering
    \includegraphics[width=0.8\textwidth]{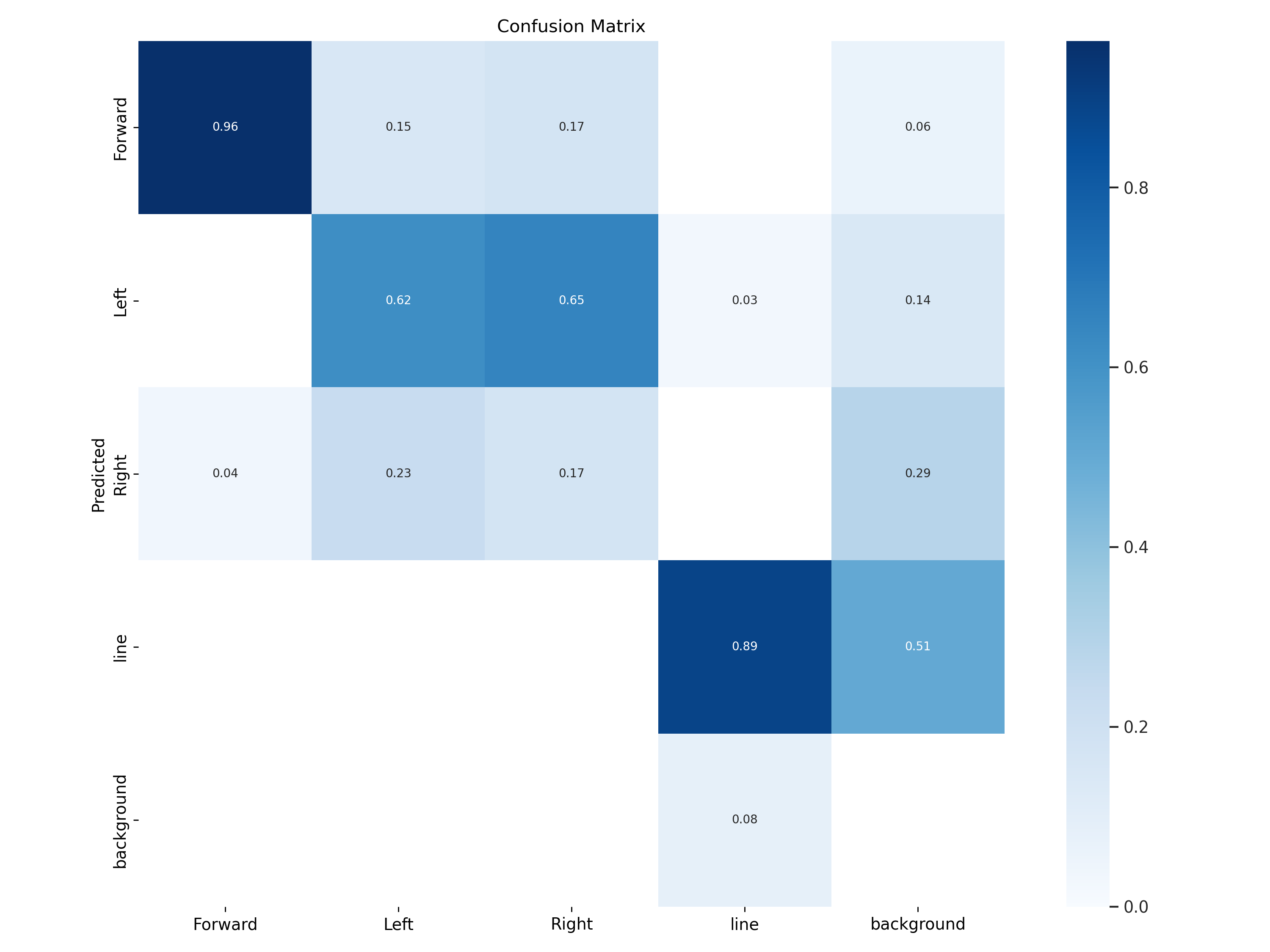}
    \caption{ The confusion matrix for YOLOv9-T's marathon model. The matrix suggests the model is frequently confusing left and right arrows due to their similar appearance.}
    \label{fig:confusion_matrix}
\end{figure}

The Marathon models performed the weakest overall, showing lower mAP and precision across all three architectures. While recall remained relatively high, the models struggled with classification accuracy. This was largely due to the similar appearance of visual markers used in the event, particularly the left and right arrows, which the models frequently confused. Analysis of the confusion matrix in Fig. \ref{fig:confusion_matrix} confirmed this as a consistent error pattern.

The primary limiting factor in model performance was dataset generalization. Given the OP3’s limited computing resources, the training dataset had to be kept small to allow for real-time inference. This constraint often led to overfitting, with models memorizing training samples rather than learning robust feature representations. Although this tradeoff limited accuracy in complex events such as Marathon, it was necessary to preserve real-time operability.

Given the superior performance in terms of frame rate and efficiency, the Tiny (YOLOv9-T) models were selected for deployment. They offered the best balance of speed and acceptable accuracy, making them ideal for the real-time requirements of the robot’s vision system. Benchmark tests, conducted with all optimizations applied and with no objects in view, confirmed that the Tiny models consistently delivered the highest frame rates under idle conditions (see Table \ref{tab:fps}). As such, the performance tests for the YOLO Vision Module will be conducted using the YOLOv9-T models. Future work may focus on leveraging transfer learning or lightweight augmentation strategies to enhance model generalization without sacrificing speed.

\begin{table}[H]
    \centering
    \renewcommand{\arraystretch}{1.5} 
    \setlength{\tabcolsep}{10pt} 
    \caption{Results for the idle FPS testing. Values were averaged over a minute with no detections for a clear baseline. }
    \begin{tabular}{|c|c|c|c|}
        \hline
        \rowcolor{purple!30}
        \multicolumn{4}{|c|}{\textbf{Idle Frame Rate Test Results}} \\
        \hline
        \rowcolor{gray!30}
        \textbf{Architecture} & \textbf{Basketball }  & \textbf{ Archery } & \textbf{Marathon}\\
        \hline
        YOLOv9-T & 7.99 FPS  & 8.01 FPS & 7.98 FPS \\
        \hline
        YOLOv9-S & 2.75 FPS &  2.75 FPS & 2.74 FPS \\
        \hline
        YOLOv9-M & 0.86 FPS & 0.86 FPS & 0.84 
        FPS \\
        \hline
    \end{tabular}
    \label{tab:fps}
\end{table}

\subsection{Experimental Setup and Procedure}

\begin{figure} 
    \centering
    \includegraphics[width=0.8\textwidth]{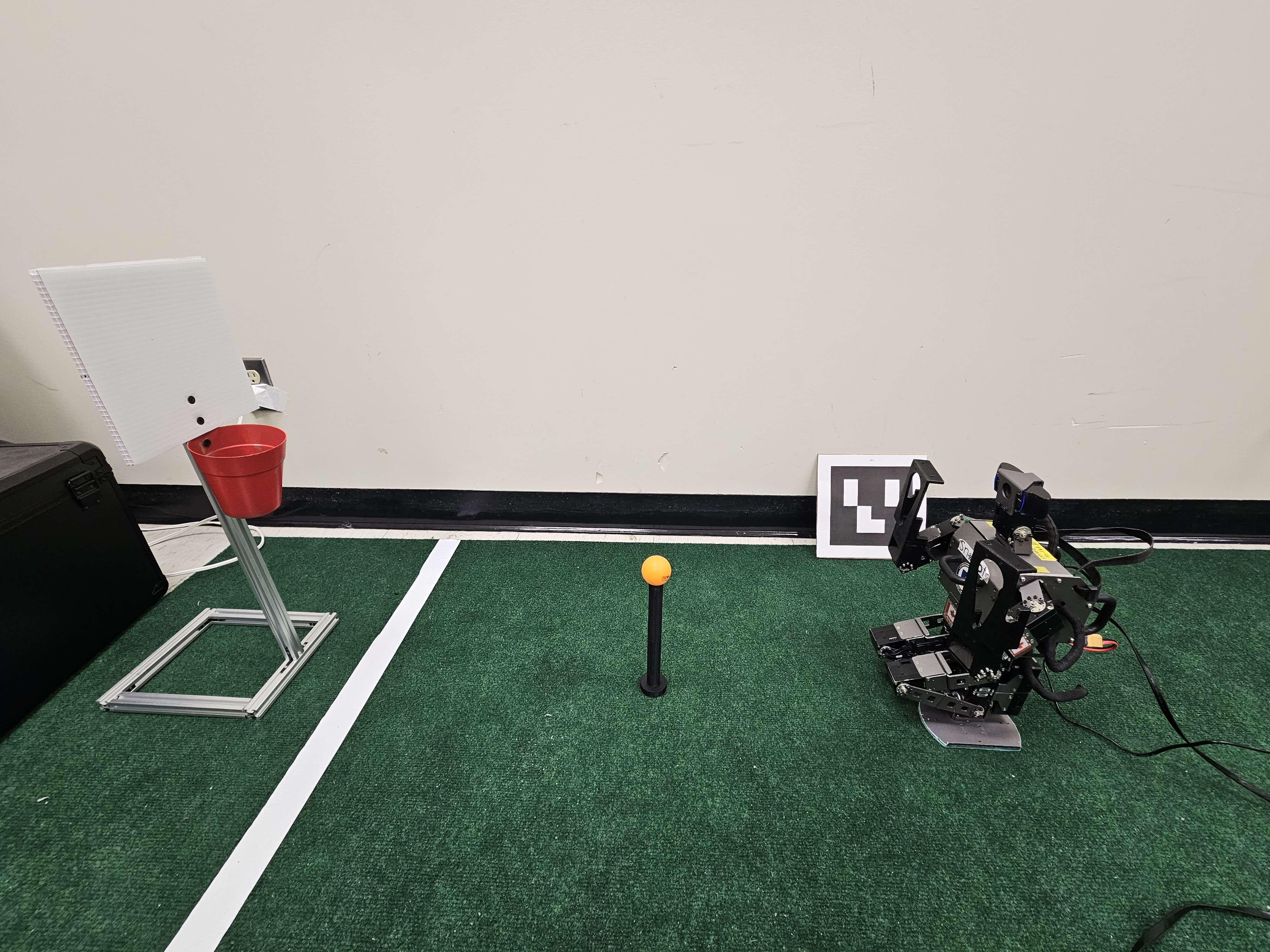}
    \caption{ All experiments, both static and dynamic, will be conducted on the LIMRL's turf field, similar to the ones utilized at the FIRA competition. This photo shows the setup for the Basketball Event.}
    \label{fig:setup}
\end{figure}

Two distinct types of experiments were conducted to evaluate the performance of the YOLOv9 module in a robotic setting: a static test and a dynamic test. These tests aim to measure the system's inference speed, real time performance, and accuracy in a controlled setting and in an active use case. Additionally, both tests will be compared to the geometric approach currently implemented on the robot. This is to serve as a baseline for assessing the vision based model's effectiveness. All tests will be conducted inside the LIMRL as shown in Fig. \ref{fig:setup}

The static test isolates the vision system’s performance by processing frames while the robot remains stationary. Key metrics include inference time and frames per second, providing a baseline for real-time feasibility and computational load. The dynamic test evaluates the system under operational conditions, where the robot uses its vision module to interact with its environment. This reveals how movement, lighting, and real-time demands affect detection and responsiveness.

Both tests are compared to a geometric baseline that uses color segmentation and rule-based logic. This comparison quantifies the trade-offs between speed, accuracy, and practicality, helping assess the benefits of the YOLO-based system in robotic applications.

\subsection{Static Experiments}

In the static experiments (see Table \ref{tab:static}), the YOLOv9 models were able to record comparable precision levels to those achieved by the existing geometric model. In the marathon event, the YOLO model has exceeded the precision levels of the geometric model. While the YOLO model made more consistent detections, it frequently confused arrow marker types, resulting in false detections. Therefore, the number of true detections achieved by the YOLO module is likely closer to the geometric model than the numbers show.

\begin{figure}
    \centering
    \includegraphics[width=0.9\linewidth]{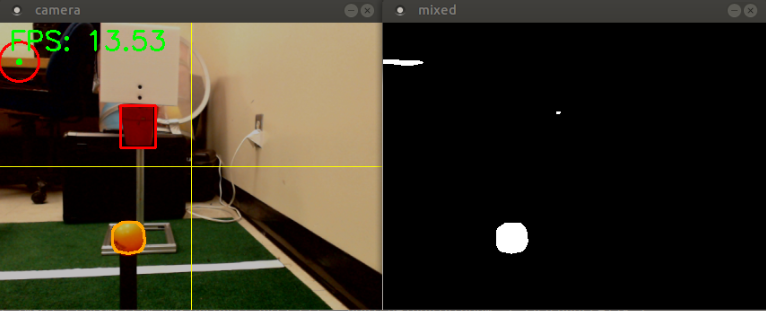}
    \caption{The figure shows an example of a case where the Geometric Node gets thrown off by background noise in the static experiment. Due to a change in lighting conditions, the node detects background objects as part of the ball, resulting in errors. The YOLO node is much more resilient to this kind of noise }
    \label{fig:noise}
\end{figure}

In terms of FPS, the geometric approach still achieves better performance, sometimes reaching speeds 50\% faster than the YOLO module. Even with the optimizations, the YOLO module still performs much slower than the geometric one, however, the YOLO module has several key advantages. Frameworks like YOLO don't require any calibration post-training, unlike the current color-segmentation-based approach, which requires a lengthy fine-tuning period to achieve good results. Secondly, through data augmentation, the YOLO module is much more resistant to environmental changes. Dataset augmentations, such as brightness or rotation adjustments, allow YOLO to achieve a level of robustness that the geometric approach cannot match. While the geometric approach often struggles with background noise (See Fig. \ref{fig:noise}), the YOLO node ignores it completely.

\begin{table}
    \centering
    \renewcommand{\arraystretch}{1.5} 
    \setlength{\tabcolsep}{8pt} 
    \caption{Results for the static testing of the new vision module. FPS measures the speed, the Precision represents the percentage of correct detections.}
    \begin{tabular}{|c|c|c|}
        \hline
        \rowcolor{yellow!30}
        \multicolumn{3}{|c|}{\textbf{Static Experiment Results}} \\
        \hline
        \textbf{Module} & \textbf{ Speed(FPS) }  & \textbf{ Precision }  \\
        \hline
        \rowcolor{gray!30}
        \multicolumn{3}{|c|}{\textbf{Basketball}} \\
        \hline
        YOLO Module & 5,92 & 91.6\% \\
        \hline
        Geometric Module & 12.99 & 91.7\%\\
        \hline
        \rowcolor{gray!30}
        \multicolumn{3}{|c|}{\textbf{Archery}} \\
        \hline
        YOLO Module & 6.22  &  92.24\%  \\
        \hline
        Geometric Module & 10.64 & 96.6\%  \\
        \hline
        \rowcolor{gray!30}
        \multicolumn{3}{|c|}{\textbf{Marathon}} \\
        \hline
        YOLO Module & 6.12  & 91.67\%* \\
        \hline
        Geometric Module & 8.21 & 50.8\%  \\
        \hline
    \end{tabular}
    \label{tab:static}
\end{table}

\subsection{Dynamic Experiments}

In terms of performance, the dynamic experiment produced results similar to those of the static experiment (see Table \ref{tab:dynamic}). The FPS is lower for both modules; however, the gap in between them remains similar. The exception to this is the marathon event, where the YOLO module can compete with the geometric module because in the marathon event, the geometric module is required to process more complex tasks to detect the arrow markers.

\begin{table}[H]
    \centering
    \renewcommand{\arraystretch}{1.5} 
    \setlength{\tabcolsep}{8pt} 
    \caption{Results for the Dynamic testing of the new vision module. FPS measures the speed, the precision represents the percentage of correct detections.}
    \begin{tabular}{|c|c|c|}
        \hline
        \rowcolor{blue!30}
        \multicolumn{3}{|c|}{\textbf{Dynamic Experiment Results}} \\
        \hline
        \textbf{Module} & \textbf{ Speed(FPS) }  & \textbf{ Precision } \\
        \hline
        \rowcolor{gray!30}
        \multicolumn{3}{|c|}{\textbf{Basketball}} \\
        \hline
        YOLO Module & 6.68 & 83,71\%  \\
        \hline
        Geometric Module & 12.30  & 50.9\% \\
        \hline
        \rowcolor{gray!30}
        \multicolumn{3}{|c|}{\textbf{Archery}} \\
        \hline
        YOLO Module & 4.41 &  90.1\% \\
        \hline
        Geometric Module & 9.24 & 93.25\% \\
        \hline
        \rowcolor{gray!30}
        \multicolumn{3}{|c|}{\textbf{Marathon}} \\
        \hline
        YOLO Module & 5.90  & 39.77\%  \\
        \hline
        Geometric Module & 8.49 & 55.5\% \\
        \hline
    \end{tabular}
    \label{tab:dynamic}
\end{table}

\begin{figure} [ht]
     \centering
     \includegraphics[width=0.5\linewidth]{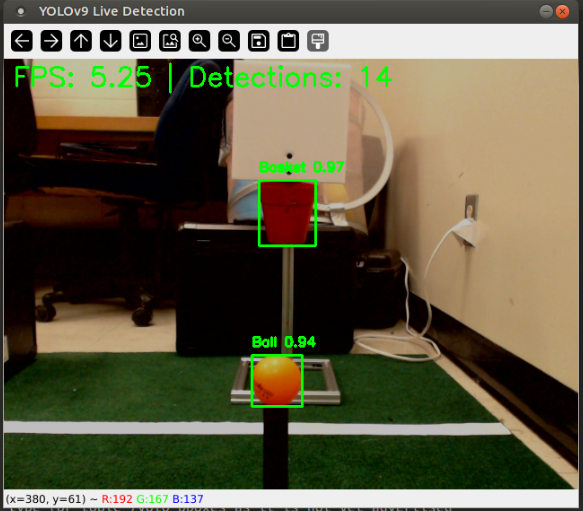}
     \caption{ YOLO is much more robust, and therefore more resilient to background noise than traditional segmented approaches }
     \label{fig:yolov9_output}
 \end{figure}

Precision varies drastically between events due to their various requirements. In basketball, the YOLO module outperformed the Geometric module. The geometric module for basketball previously used contour detection, and it struggles with detecting spherical objects due to variances in lighting conditions. The ball detection is set up into two different segmentations and combined to account for the light; however, the approach remains cumbersome. The YOLO module is much better equipped to deal with this variability. In Archery, both modules performed similarly, and in Marathon, the Geometric module outperformed the YOLO module. Marathon requires the detection of arrow markers, and the YOLO module has a hard time differentiating between marker types. Figure~\ref{fig:yolov9_output} shows the output from the basketball event.

\subsection{Discussion}
Throughout static testing, the geometric module consistently outperformed YOLO in frames per second (FPS), averaging 50\% higher across all scenarios due to its lower computational cost. However, its advantage diminished in complex tasks. YOLO maintained a stable frame rate across tests, suggesting optimization potential through pruning or hardware acceleration. A breakdown of per-frame processing time could clarify where each model incurs computational costs.

Precision results were more event-dependent. Basketball and Archery yielded comparable scores, while Marathon showed YOLO achieving higher detection rates—but with frequent misclassifications between similar arrow markers, diminishing its effective accuracy. This suggests YOLO's confidence scores may not always reflect real-world reliability. In contrast, the geometric model’s consistency made it less prone to such false positives, especially with repetitive object shapes. Further tests under distortions like motion blur could assess robustness more thoroughly.

In dynamic tests, with movement and real-time decision-making, FPS trends remained consistent: the geometric module led in speed but with a narrowing gap in Marathon due to detection complexity. Event-specific challenges influenced performance. YOLO excelled in Basketball, handling lighting changes and background clutter better than the geometric module, which struggled with occlusions. Marathon again exposed YOLO’s weakness in distinguishing similar markers, while the geometric module maintained higher precision. Archery offered balanced conditions where both methods performed well.

Precision varied significantly by event. YOLO outperformed in Basketball due to its adaptability to visual noise. In Archery, both modules achieved high accuracy, with the geometric node slightly ahead. In Marathon, YOLO’s precision dropped sharply due to misclassification. These outcomes emphasize that the optimal module depends on task context: geometric methods suit controlled environments or speed-critical applications, while YOLO thrives in dynamic, unpredictable scenes with adequate resources.

Overall, the results highlight a trade-off: the geometric module is fast and effective in simple tasks but lacks adaptability, while YOLO handles complex scenes at the cost of computational load. A hybrid solution—YOLO for detection, geometric for refinement—could combine their strengths. While deep learning tools continue to evolve, both methods remain valid, with choice driven by task demands and hardware limitations.

\section{Conclusions}

\subsection{Summary of Findings}

This research successfully demonstrated the feasibility of deploying a YOLO-based vision module on a humanoid robot, overcoming computational challenges through targeted optimizations. By conducting both static and dynamic experiments, the study provides a comprehensive evaluation of the trade-offs between deep-learning-based object detection and the traditional geometric approach. The results do not show that one of these modules is superior, instead, they highlight the strengths and limitations of each method, offering insights for robotic applications.

The static experiments confirmed that YOLO can achieve high detection accuracy, often matching the geometric approach in precision. However, the geometric module remains considerably faster, outperforming YOLO in terms of computational efficiency and processing frames at nearly twice the speed in some scenarios. Although expected, this efficiency gap highlights the need for further optimization, especially given OP3's hardware limitations. The dynamic experiments further emphasized this trade-off. While YOLO showed much better robustness, its real-time performance suffered further. The geometric approach retained its speed advantage, however, it struggled with certain complex object types. This indicates that while deep learning models introduce significant computation overhead, their ability to generalize across varying conditions can significantly enhance real-world usability

\subsection{Limitations and Challenges}

Despite the system's success in achieving real-time detection under controlled conditions, several limitations must be acknowledged. Firstly, the OP3 platform lacks a dedicated GPU, requiring extensive optimizations to maintain acceptable performance. Even with the quantization and model conversions, inference speed remained a considerable bottleneck in certain high-demand scenarios. Secondly, due to the necessity of maintaining a small dataset, the models exhibited signs of overfitting, particularly in tasks requiring difficult object differentiation. While the system performed well in controlled settings, variations in lighting and background clutter still slightly impacted detection accuracy in real-world scenarios. Even with augmentations, the change in conditions can be so drastic that the need for more adaptive training data augmentation strategies becomes apparent. Addressing these challenges will require a combination of hardware improvements, additional software optimizations, and refinements to the training process to enhance generalization across diverse operating conditions.

\subsection{Future Work}
This study opens several directions for future research. One key area is hardware acceleration. Leveraging edge computing devices like Intel Neural Compute Sticks or Raspberry Pi AI accelerators could offload inference and reduce computational strain. In networked environments, cloud-based inference or GPU/FPGA-optimized implementations may offer significant performance gains, especially where low-latency is critical.

Expanding the dataset to include diverse environmental conditions and using transfer learning can also enhance model generalization. A larger or more varied dataset—potentially aided by synthetic data generation—would help address the overfitting seen with small samples.

Another promising avenue is multi-modal sensing. Combining YOLOv9 with depth sensors or LiDAR could improve object localization, especially in occluded or noisy scenes. Sensor fusion techniques like Kalman Filtering may reduce uncertainty and enhance detection robustness.

Finally, adaptive inference strategies—such as dynamic frame skipping, resolution scaling, or model switching based on system load—could optimize performance. Reinforcement learning or self-supervised methods could allow models to adapt to real-world conditions without full retraining.

In conclusion, this work demonstrates the viability of a YOLOv9-based vision system on a humanoid robot, overcoming hardware constraints through targeted optimizations. It offers a flexible foundation for integrating deep learning in robotics, with potential across fields like autonomous navigation, assistive tech, and industrial automation.

\bibliographystyle{unsrtnat}
\bibliography{references}  






\end{document}